# An economically-consistent discrete choice model with flexible utility specification based on artificial neural networks

José Ignacio Hernández [a,*], Niek Mouter [b, c], Sander van Cranenburgh [b]

[a] Center of Economics for Sustainable Development (CEDES), Faculty of Economics and Government, Universidad San Sebastián. Lientur 1457, Concepción, Chile.

[b] Transport and Logistics Section, Department of Engineering Systems and Services, Faculty of Technology, Policy and Management, Delft University of Technology. Jaffalaan 5, 2628BX Delft, The Netherlands.

[c] Populytics. Research Agency. Frambozenweg 139, 2321KA Leiden, The Netherlands.

[*] Corresponding Author. E-mail: Jose.Hernandez@uss.cl

**Abstract**

Random utility maximisation (RUM) models are one of the cornerstones of discrete choice modelling. However, specifying the utility function of RUM models is not straightforward and has a considerable impact on the resulting interpretable outcomes and welfare measures. In this paper, we propose a new discrete choice model based on artificial neural networks (ANNs) named "Alternative-Specific and Shared weights Neural Network (ASS-NN), which provides a further balance between flexible utility approximation from the data and consistency with two assumptions: RUM theory and fungibility of money (i.e., "one euro is one euro"). Therefore, the ASS-NN can derive economically-consistent outcomes, such as marginal utilities or willingness to pay, without explicitly specifying the utility functional form. Using a Monte Carlo experiment and empirical data from the *Swissmetro* dataset, we show that ASS-NN outperforms (in terms of goodness of fit) conventional multinomial logit (MNL) models under different utility specifications. Furthermore, we show how the ASS-NN is used to derive marginal utilities and willingness to pay measures.

## 1    Introduction

Random utility maximisation (RUM) models (McFadden, 1974) are one of the cornerstones of discrete choice modelling. RUM models provide a framework to analyse travel demand (Ben-Akiva & Lerman, 1985; Ben-Akiva & Bierlaire, 2003). The strength of RUM models relies on their interpretability and connection with economic theory (Small & Rosen, 1981). Notably, the estimates of RUM models can inform the analyst about individual preferences for attribute changes, substitution rates and willingness to pay. This property of RUM models makes them a particularly insightful approach for transport appraisal.

However, a key challenge of RUM models is the specification of the utility function. Conventionally, specifying the utility of a RUM model concerns a trial-and-error process, where the analyst estimates several competing models (with different functional specifications), based on prior knowledge (e.g., findings from previous studies, economic theory). The analyst selects the final specification based on, on the one hand, behavioural intuition (i.e., the size and magnitude of the estimated parameters make sense) and, on the other hand, goodness-of-fit or information criteria. Nevertheless, the true utility functional form is not known but assumed *a priori* by the analyst. Furthermore, the selected utility specification has a considerable impact on the derived interpretable measures, such as the estimated willingness to pay (Torres et al., 2011; van der Pol et al., 2014). In consequence, the selected utility specification is not a trivial choice, considering the relevance of such measures for policymaking and appraisal.

An alternative that can help to circumvent this challenge of RUM models is using machine learning (ML) models. ML models are methods aimed to learn patterns and/or approximate mathematical functions directly from the data. In the last years, ML models have been increasingly adopted in the discrete choice modelling field (van Cranenburgh et al., 2022). ML models differ from discrete choice models (DCMs) in their modelling paradigms. DCMs are theory-driven, in the sense that the analyst assumes the underlying data-generating process (DGP), and the goal is to find the model parameters that better describe such model, given data. Unlike DCMs, ML models adopt a data-driven paradigm under the guiding principle that the true DGP is unknown and complex, but it can be uncovered from the data. This difference in paradigms allows ML models to reach higher performance than their theory-driven counterparts for predictive tasks (Wang, Mo, et al., 2021).

Among specific ML models, Artificial Neural Networks (ANNs) have gained considerable ground in the discrete choice modelling field (e.g., Alwosheel et al., 2018; Sifringer et al., 2020; van Cranenburgh & Alwosheel, 2019; Wang, Mo, et al., 2020; Wang, Wang, et al., 2021). ANNs are ML models loosely based on the structure of brains, aimed to approximate mathematical functions from data. In an ANN, the underlying DGP is modelled as a set of layers and nodes interconnected to each other. This allows ANNs to model complex interactions between input variables (covariates) without the need to be specified *a priori* by the analyst. Notably, ANNs can be structured to build discrete choice models with a flexible utility function (Bentz & Merunka, 2000), which allows accounting for interactions and nonlinear effects that the analyst could overlook and, therefore, overcoming the limitations of manually specifying a specific utility function.

Despite their strengths, ANNs provide limited information of behavioural and economic interest without further intervention. This is because the parameters of ANNs lack interpretation, as a difference from conventional discrete choice models. To overcome this limitation, several works have proposed alternative structures that restrict part of the ANN to increase its interpretability (Han et al., 2022; Sifringer et al., 2020; Wong & Farooq, 2021). This strategy, however, involves a trade-off between having a flexible utility function, i.e., that can capture interactions and non-linearities without being specified *a priori*, and consistency with RUM and economic theory to derive measures that can be used for welfare analysis. On the one hand, an ANN with the highest flexibility to approximate utility functions (e.g., an ANN with no intervention) provides outcomes that may violate consistency with RUM theory and, therefore, the connection between their derived welfare measures and economic theory cannot be guaranteed. On the other hand, an ANN with a high level of intervention would provide interpretable outcomes that satisfy RUM and economic assumptions but at the expense of having a utility specification that is not flexible enough to identify interactions or nonlinear effects from the data. To balance these trade-offs, the analyst must provide a structure that provides enough flexibility to the ANN to approximate utility functions and, at the same time, satisfies consistency with RUM and economic assumptions that guarantee to derive meaningful interpretable outcomes and welfare measures.

In this paper, we propose the "Alternative-Specific and Shared weights Neural Network" (ASS-NN), a discrete choice model based on ANNs that incorporates domain knowledge to guarantee consistency with RUM and economic theory. The ASS-NN is built upon the "Alternative-Specific Utility Deep Neural Network" (ASU-DNN), proposed by Wang, Mo, et al. (2020). Both models feature alternative-specific utility functions that are approximated from the data. Our proposed model, in addition, postulates "fungibility of money", also known as "one euro is one euro". Fungibility of money refers to the notion that money can be spent in different goods (alternatives) interchangeably. As a result of this assumption, the marginal utility of costs for the same individual must be equal across alternatives of the same cost level. In addition, we discuss that the alternative-specific utility structure of both the ASU-DNN and the ASS-NN are consistent with RUM (Hess et al., 2018) and, in consequence, the outcomes and welfare measures obtained from the ASS-NN are better connected with economic theory. To incorporate fungibility of money, the cost-dependent utility of the ASS-NN is modelled with separate sets of hidden layers with shared weights, which forces equal marginal utility of costs across alternatives for a same individual and same cost level. The trained ASS-NN can be used to estimate the marginal utility of attribute increases, marginal rates of substitution, and willingness to pay for attribute changes, e.g., the value of travel time (VTT).

We show the use of the ASS-NN using a Monte Carlo experiment and empirical data from the *Swissmetro* dataset (Bierlaire et al., 2001). The Swissmetro dataset is a mode choice experiment that is widely known by the transportation research community, and it has been previously used as benchmark data in other ANN-based discrete choice models proposed in the literature (e.g., Sifringer et al., 2020). We first conduct a Monte Carlo experiment with pseudo-synthetic choices generated from the Swissmetro dataset to show that the ASS-NN succeeds in approximating the true utility function from the data under different utility specifications, as well as in recovering the marginal utility of attribute increases and willingness to pay. Then, we train the ASS-NN with empirical data to approximate the marginal utilities and willingness to pay, and we compare these outcomes with those from the ASU-DNN and conventional multinomial logit (MNL) models under different utility specifications. To allow researchers to replicate our work and encourage open science, the code and data used in this paper are published in a Git repository: https://github.com/ighdez/ass_nn_paper.

The remainder of this paper is as follows. Section 2 describes the methodology and how the ASS-NN is implemented. Section 3 presents the setting and results of the Monte Carlo experiment. Section 4 presents the empirical data and results. Finally, section 5 provides a discussion and conclusion.

## 2 Methodology

### 2.1 Theoretical models

The RUM model is a theoretical framework to describe individual choice behaviour based on the notion that decision-makers seek to maximise their utility from a set of discrete goods. In this section, we proceed to formalise a general RUM model that can be approximated with the ASS-NN.

Let $n$ be a decision-maker that perceives utility from the consumption of $J$ mutually-exclusive goods. Each alternative $j$ is characterised by $K$ observable attributes (Lancaster, 1966). For a given alternative $j$ faced by decision-maker $n$, let $X_{nj} = \{x_{n11}, \dots, x_{n1K}\}$ be the vector of non-cost attributes of such alternative, and $c_{nj}$ be the cost value. Then, the utility of each good $U_{nj}$ perceived by decision-maker $n$ for alternative $j$ is a function of such observable characteristics plus a stochastic error term, as defined by equation (1):

$$U_{nj} = V_n(X_{n1}, X_{n2}, \dots, X_{nJ}, c_{n1}, c_{n2}, \dots, c_{nJ}, w) + \varepsilon_{nj}, \qquad (1)$$

where $V_n$ is a function that depends on observed attributes and costs, $w$ is a vector of weights to be estimated and $\varepsilon_{nj}$ is a stochastic error term. The model described in equation (1) is a general discrete choice model without a specific utility functional form that can be approximated with a fully-connected ANN (Bentz & Merunka, 2000).

However, a limitation of this model is that is not consistent with RUM theory. As discussed by Hess et al. (2018), a RUM-consistent model must satisfy two conditions: regularity and transitivity. Regularity refers that adding a new alternative to the choice set should not increase the choice probability of the other alternatives. Transitivity states that if an alternative A is preferred over B, and B over C, then alternative A is preferred over C. The model of equation (1) does not exhibit regularity, since the attributes of one alternative can affect the utility of other alternative(s). Thus, if an alternative is added to the choice set, and the attributes of such alternative affects the utility of other alternatives in such a way that their choice probabilities increase, then regularity is violated, and the model is not consistent with RUM.

The ASU-DNN model (Wang, Mo et al., 2020) overcomes this limitation. The ASU-DNN is an ANN-based model that features alternative-specific utility functions that are approximated from the data. Formally, the ASU-DNN specifies the utility as in equation (2):

$$U_{nj} = V_{nj}(X_{nj}, c_{nj}, w_j) + \varepsilon_{nj}, \qquad (2)$$

where $V_{nj}$ is the observed utility of alternative $j$ and $w_j$ is an alternative-specific vector of weights to be estimated for alternative $j$. As a difference with the model of equation (1), the ASU-DNN model is consistent with RUM since $V_{nj}$ only depends of its corresponding attributes and costs.

However, the ASU-DNN does not restrict the cost-dependent utility function to have the same form across different alternatives, which implies that the fungibility assumption does not hold. Behaviourally speaking, in the ASU-DNN, a decision-maker could value one euro spent on a specific alternative in an intrinsically different way than the same euro spent on another alternative. However, a key requirement for deriving meaningful welfare measures is that money is a perfect substitute of itself. Thus, under the ASU-DNN, we cannot ensure that euros are interchangeable across different alternatives of the choice set, which makes welfare measures unfeasible to compare.

To remedy this issue, we propose a to modify the utility function of equation (2) to incorporate the fungibility of money assumption, such as in equation (3)

$$U_{nj} = f_j(X_{nj}; w_j) + g_j(c_{nj}; \bar{w}_c) + \varepsilon_{nj}, \qquad (3)$$

where $f_j(\cdot)$ is an alternative-specific utility function of alternative $j$ that depends only on the non-cost attributes of the same alternative and $g_j(\cdot)$ is the utility function of alternative $j$ that depends only on the costs of the same alternative. The vectors $w_j$ and $\bar{w}_c$ are weights (parameters) to be estimated, which define the shape of $f_j$ and $g_j$, respectively. The vectors $w_j$ are alternative-specific, meaning they are independent across alternatives, whereas the vector $\bar{w}_c$ is shared across alternatives.

This model is consistent with RUM since the utility functions are alternative-specific and only depend on their corresponding attributes (in the same way as the ASU-DNN). In addition, this model is also consistent with the fungibility of money assumption since the cost-dependent utility functions only depend on their own attributes, but the weights are shared across alternatives. Therefore, the cost-dependent utility functions share the same shape, leading to the same marginal utility of costs across different alternatives for a given cost value.

### 2.2 *The alternative-specific and shared weights neural network (ASS-NN)*

The ASS-NN is an ANN structure aimed to approximate the utility function of equation (3). Figure 1 illustrates an example of the ASS-NN for a choice situation composed of three alternatives and three attributes per alternative, namely travel cost, travel time and an additional attribute that is present only in the first two alternatives. The travel costs of each alternative are modelled with an alternative-specific hidden layer with shared weights, named as "shared layer". The shared layer only receives the information from the costs of its correspondent alternative and translates it into a single utility value. The non-cost attributes, i.e. the travel time and the additional attribute for this example, are modelled with regular alternative-specific hidden layers (i.e., with independent weights), named as "alternative-specific-utility (ASU) layers". For each alternative, the utility values coming from the shared ASU layers are summed to form a single utility associated to the alternative. Optionally, the utility of each alternative can incorporate bias nodes that mimic the alternative-specific constants of a discrete choice model. Finally, the utility values are transformed to choice probabilities using a Softmax function that guarantees that the probabilities sum up to one.

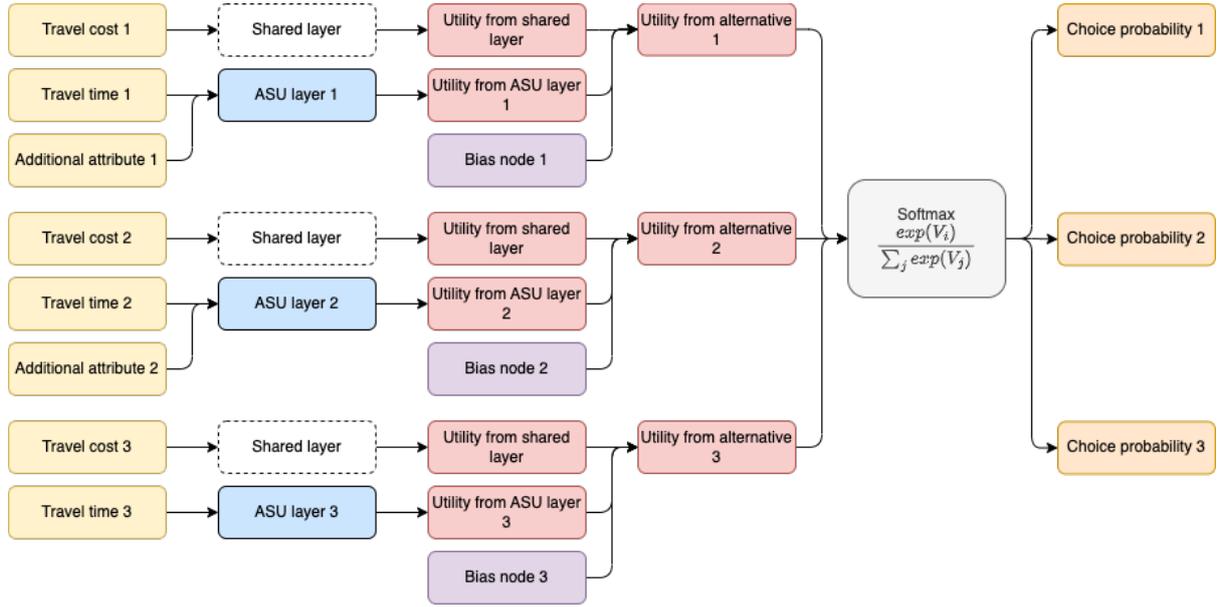

Figure 1: Illustration of the ASS-NN for a 3-alternatives choice situation

Given data with $N$ observations, the ASS-NN is trained by finding the weights that minimise the categorical cross-entropy (CE) function described by equation (3):

$$CE(y, p; w) = \left(\frac{1}{N}\right) \cdot \sum_{n=1}^{N} \sum_{j=1}^{J} \ln(p_{nj}) \cdot y_{nj}. \tag{3}$$

where $p_{nj}$ is the probability of choosing alternative $j$ by decision-maker $n$, and $y_{nj}$ is a binary variable that is equal to one when alternative $j$ is chosen by the decision-maker. The CE function is an averaged version of the log-likelihood function of discrete choice models.

## 2.3 Implementation and training of the ASS-NN

The ASS-NN is implemented in three steps. In the first step, we prepare the data in a compatible format, and we split it into a training (estimation) and out-of-sample testing (prediction) dataset. The second step consists of finding the optimal hyperparameters. The last step consists of training the ASS-NN and deriving outcomes from it using simulation, namely the marginal utility of attribute increases and the VTT (Small, 2012), a specific form of the willingness to pay for attribute changes based on marginal rates of substitution. Below, we detail each of these three steps.

### 2.3.1 Step 1: Data preparation

Table 1 describes the basic dataset structure for the ASS-NN. The data is arranged in "wide" format, the standard format in choice modelling statistical packages (e.g., Biogeme, Apollo). Each row represents a single choice situation, while each column represents the variables of such choice situations. The minimum variable requirements are 1) an integer variable that identifies the selected alternative (Choice), 2) the cost attributes that are modelled with shared layers (TC1 and TC2), and 3) the non-cost attributes that are modelled with ASU layers, such as travel time (TT1 and TT2). The cost variables that are modelled with shared weights must be present in all alternatives by construction, while this is not required for the non-cost attributes that are modelled with ASU layers.

Table 1: Data format for the ASS-NN

| Respondent ID | Choice | TT1 | TC1 | TT2 | TC2 |
|---|---|---|---|---|---|
| 1 | 2 | 50 | 15 | 45 | 18 |
| 2 | 1 | 65 | 10 | 70 | 8 |
| 3 | 1 | 50 | 10 | 58 | 8 |
| … | … | … | … | … | … |

Before training the ASS-NN, data is normalised to avoid numerical overflow issues during the optimisation of the CE function. We use so-called Min-Max normalisation, in which each variable is scaled between zero and one using the minimum and maximum values of the variable as bounds for the normalisation. Specifically, the Min-Max normalisation applies the transformation detailed in equation (4):

$$x_{scaled} = \frac{x - \min(x)}{\max(x) - \min(x)}, \qquad (4)$$

where $x_{scaled}$ is the scaled value of the variable $x$. For the cost variables, the normalisation is done considering the minimum and maximum values of all cost variables as bounds, since such variables are modelled with hidden layers with shared weights.

After normalising, data is randomly split in a training sample (80% of the data) used to train the ASS-NN, and a test sample (the remaining 20% of the data) used for out-of-sample prediction. This is done because ANNs, in general, are prone to overfit, which hinders the ability of the model to provide generalisable measures for data points that were not used for training. In addition, using split samples prevent data-leakage issues, i.e., when a model learns from the test sample, which may provide overly optimistic predictions (Hillel et al., 2021). We use a stratified random sampling of the data, using the choice variable for stratification to ensure that both training and test samples keep the same observed market shares.

### 2.3.2 Step 2: Optimal hyperparameters search

Before training the ASS-NN, it is necessary to determine the optimal hyperparameters. Specifically, we focus on the network topology and the activation functions. On the one hand, the network topology is the number of hidden layers and nodes per layer and determines the ability of the ASS-ANN to learn from the data. A too-simple network topology (i.e., too few hidden layers/nodes) will underfit the training sample, and the ASS-NN will not be able to learn relevant interactions from the data. A too-complex network topology will overfit the training sample, hindering their generalisation outside the domain of the training data. By finding the optimal network topology, we aim for a model that maximises out-of-sample predictive performance. On the other hand, the activation functions are transformations located on each hidden node and their role is to convert the information from each preceding hidden layer and pass the transformed information to the subsequent layers. Non-linear activation functions are defined to allow the ASS-NN to identify nonlinear effects from the data.

We use a grid search procedure for finding the optimal hyperparameters. Table 2 summarises the candidate hyperparameters for the hidden layers, nodes per layer and activation functions. A set of ASS-NNs for each possible combination of hyperparameters is trained 100 times each. Repeated training is performed because ANNs are overspecified models which are not necessarily globally concave/convex, making them prone to get stuck in local minima. Performing repeated training mitigates that risk, as the prediction performance of networks with poor solutions can be compensated by networks with a better performance. On each training repetition, 20% of the last observations of the training sample are taken apart, and the ASS-NN is trained using the remaining 80%. This excluded sample, known as validation data, is used to calculate a validation cross-entropy on each training epoch (iteration). The training process stops when the validation cross entropy does not improve for 6 consecutive epochs.

Table 2: Hyperparameter specifications

| Parameter | Values |
| --- | --- |
| Hidden layers | 1, 2 |
| Hidden nodes per layer | 1 layer: 5, 6, 7, 8, 9, 10, 15, 20, 30 |
|  | 2 layers: 5, 10, 20, 30 |
| Activation functions | ReLU, Hyperbolic tangent (tanh) |

For each training repetition, we use the test sample to compute the log-likelihood (i.e., the unaveraged version of the CE function) and the Rho-squared. The Rho-squared compares the predictive ability of a discrete choice model with a random sampling of the choice probabilities (Mokhtarian, 2016), as defined in equation (5):

$$\rho^2 = 1 - \frac{LL_{test}}{N_{test} \cdot ln\,(1/J)}, \tag{5}$$

where $LL_{test}$ is the log-likelihood value obtained from the test sample, $N_{test}$ is the test sample size, and $J$ is the number of alternatives. Higher values indicate that the model reaches a better prediction performance than mere random chance. Thus, the optimal hyperparameters correspond to those that result in the model that minimises the test log-likelihood and Rho-squared.

2.3.3 Step 3: Training and simulation of outcomes

After the optimal hyperparameters are identified, the ASS-NN is trained 100 times using the training dataset to mitigate the possibility of predicting on a single, poor-performing network. The predictions of each network are averaged across the 100 training repetitions. Same as in step 2, on each training repetition, the last 20% of the training data is taken apart to calculate the validation CE on each epoch and end the training process if no further improvements of this metric are found for six consecutive epochs.

The trained ASS-NN is used to derive the marginal utility of attribute increases and the VTT using simulation, following the approach of Wang, Wang, et al. (2020). Firstly, for a given decision-maker $n$ and alternative $j$, the marginal utility (MU) of such alternative with respect to increments of the attribute $k$ is given by equation (6):

$$MU_{njk} = \partial \widehat{V}_j/\partial x_{njk} = \partial \widehat{f}_j/\partial x_{njk} + \partial \widehat{g}_j/\partial x_{njk}, \qquad (6)$$

where $\widehat{f}_j$ and $\widehat{g}_j$ are the approximated utility functions that depend on the non-cost and cost attributes, respectively. When $x_{njk}$ is non-cost attribute, the $\partial \widehat{g}_j/\partial x_{njk} = 0$. In contrast, if $x_{njk}$ is the cost, $\widehat{f}_j/\partial x_{njk} = 0$. The MU for attribute increases is computed per decision-maker since $\widehat{f}_j$ and $\widehat{g}_j$ are functions of the corresponding attribute and, therefore, their associated MUs depend on the attribute values.

The MU for attribute increases provide behavioural information about the decision-makers' preferences for increases in specific attributes. For a decision-maker $n$, if $MU_{njk} > 0$, then increases of the attribute $k$ in alternative $j$ are preferred. Conversely $MU_{njk} < 0$, then increases of the attribute $k$ in alternative $j$ are not preferred by decision-maker $n$. If $MU_{njk} = 0$, decision-maker $n$ utility for alternative $j$ is not affected for changes in attribute $k$. This interpretation is similar as the estimated taste parameters of a linear-in-parameters MNL model, but in an individual-specific way.

The VTT is constructed as the marginal rate of substitution (MRS) between two attributes. Mathematically, the MRS between attributes $k$ and $l$ for decision-maker $n$ is defined by the ratio of marginal utilities, as shown in equation (7):

$$MRS_n^{kl} = MU_{njk}/MU_{njl} \qquad (7)$$

The MRS between two attributes provides information about the extent that a given decision-maker is willing to substitute attributes $k$ and $l$ to keep their utility for alternative $j$ without changes.

When the denominator of equation (7) is the MU of cost, the MRS becomes the VTT, which is the willingness-to-pay for reductions in travel time, in terms of travel costs (Small, 2012). Similar willingness-to-pay measures can be derived from the VTT expression. For instance, the ratio between the MU of travel headway (i.e., the time distance between train services) and the MU of costs is known as the Value of Waiting Time (VoWT), which is the willingness to pay for increasing the frequency of public transport services, in terms of travel costs. Both the VTT and VoWT are defined by equations (8) and (9):

$$VTT_n = MU^{TT}{}_n/MU^{TC}{}_n \qquad (8)$$

$$VoWT_n = MU^{FREQ}{}_n/MU^{TC}{}_n \qquad (9)$$

## 3 Monte Carlo analysis

To show the extent that the ASS-NN learns the utility from the data, we conducted a Monte Carlo analysis. Specifically, we generate pseudo-synthetic data based on the stated preference (SP) part of the Swissmetro dataset (Bierlaire et al., 2001), hereafter the Swissmetro data. The Swissmetro data is a mode choice experiment widely known in the transportation research community and has been used before as a benchmark dataset for ANN-based discrete choice models (e.g., Sifringer et al., 2020). The Swissmetro dataset was carried out in 1998 in Switzerland to elicit travellers' preferences for the Swissmetro, an innovative rail-based rapid transport mode. Respondents were presented with hypothetical mode choice situations based on their current trip offering three alternatives: train, Swissmetro or car. Each alternative varied in terms of their travel time in minutes, travel cost in Swiss Francs (CHF) and headway (for train and Swissmetro) in minutes between each service. After pre-processing the data and cleaning choice situations with less than 3 alternatives, the experimental design to generate pseudo-synthetic data comprises 9,036 combinations of attributes from 1,858 individuals.

*3.1 Pseudo-synthetic data generation*

Pseudo-synthetic data is generated by using the experimental design of the Swissmetro dataset (i.e., the attributes of each alternative) to generate synthetic choices, based in RUM models under different utility function specifications and "true" parameters (Garrow et al., 2010). As we know the "true" DGP *a priori*, we can contrast the outcomes of the ASS-NN, namely goodness-of-fit measures, marginal utilities and willingness to pay measures, with the "true" outcomes.

We generate two pseudo-synthetic datasets, using the travel time and travel cost of each alternative from the Swissmetro dataset and two different model specifications of the utility function that are commonly observed on empirical applications of discrete choice models. Table 3 summarises the specification of both datasets. The first dataset is generated using a linear-in-parameters utility function. This specification leads to marginal utilities equal to the corresponding parameters associated to travel time and travel cost, respectively. Furthermore, the marginal utilities are constant across different modes, which in turn determines that the VTT is the same for all modes. The second dataset follows a log-linear utility function. To avoid numerical overflow of the natural logarithm when the travel cost is zero (i.e., when a respondent holds an annual discount card), we added a constant of 0.1 to all attributes. Under the log-linear utility specification, the marginal utilities and VTT depend on the current travel time and cost of each respondent, which implies that the VTT could differ across different travel modes and respondents.

Table 3: Model specification of pseudo-synthetic datasets

| Name | Utility function | True parameters | Marginal utility | VTT |
|---|---|---|---|---|
| Dataset 1 (Linear) | $V_j = \beta_{TC} \cdot TC_j + \beta_{TT} \cdot TT_j$ | $\beta_{TC} = -2$ $\beta_{TT} = -3$ | $MU_{TC} = -2$ $MU_{TT} = -3$ | $VTT = 3/2$ (for all alternatives) |
| Dataset 2 (Log-linear) | $V_j = \beta_{TC} \cdot \ln(TC_j + 0.1) + \beta_{TT} \cdot \ln(TT_j + 0.1)$ | $\beta_{TC} = -3$ $\beta_{TT} = -5$ | $MU_{TC} = -2/(TC_j + 0.1)$ $MU_{TT} = -3/(TT_j + 0.1)$ | $VTT = \frac{2}{3} \cdot \frac{TC_j + 0.1}{TT_j + 0.1}$ |

We expect the ASS-NN to recover the marginal utility of travel time and cost increases, as well as the VTT values. Furthermore, we expect that the average values of the marginal utility and VTT are close to the corresponding true values. In addition, we expect the ASS-NN outperforms a MNL model with a linear-in-parameters utility function as the DGP departs from a linear model.

*3.2 Results of the Monte Carlo Analysis*

Table 4 summarises the training results across the 100 repetitions of the ASS-NN, trained on each pseudo-synthetic dataset. We present the log-likelihood evaluated in the full, train and test samples and the Rho-squared evaluated in the test sample. These values are contrasted with the true goodness-of-fit measures and those obtained from a MNL model with a linear utility function. In addition, we present the optimal hyperparameters of the ASS-NN for each dataset. The measures of the linear MNL model are included to show the extent that a misspecified model leads to poor predictive performance when the model assumptions are not aligned with the true DGP.

Table 4: Training results of the ASS-NN against true values and from a linear MNL model

|  | Dataset 1 (linear) | | | Dataset 2 (log-linear) | | |
| --- | --- | --- | --- | --- | --- | --- |
|  | True value | Linear MNL | ASS-NN | True value | Linear MNL | ASS-NN |
| Log-likelihood (full sample) | -5,807.57 | -5,807.07 | **-5,809.23** | -4,535.97 | -5,056.09 | **-4,562.64** |
| Log-likelihood (training sample) | -4,621.36 | -4,620.88 | **-4,623.59** | -3,618.93 | -4,041.19 | **-3,641.88** |
| Log-likelihood (test sample) | -1,186.20 | -1,186.19 | **-1,186.00** | -917.04 | -1,014.90 | **-920.76** |
| Rho-squared (test sample) | 0.40 | 0.40 | **0.40** | 0.54 | 0.49 | **0.54** |
| Estimation/training time (secs.) | - | < 1 s. (total) | **6.73s./train** | - | < 1 s. (total) | **13.04/train** |
| Optimal number of hidden layers | - | - | **1 layer** | - | - | **2 layers** |
| Optimal number of nodes per layer | - | - | **15 nodes** | - | - | **10 nodes** |
| Activation function | - | - | **tanh** | - | - | **tanh** |

Note: The goodness-of-fit metrics of the ASS-NN are the averaged values across the 100 repetitions, per dataset.

We observe that the ASS-NN reaches a goodness-of-fit close to the true values, in all samples and datasets, suggesting this model succeeds in approximating the utility function from the data. Compared with a linear MNL model, the ASS-NN reaches a negligibly lower predictive performance than the choice model, i.e., lower log-likelihood and Rho-squared, when the data is generated with a linear-in-parameters utility function (dataset 1). This result is expected, as the linear MNL is correctly specified for this pseudo-synthetic dataset. However, when the data is generated with a log-linear utility function (dataset 2), the ASS-NN outperforms the linear MNL model in terms of goodness-of-fit measures, e.g., a Rho-squared in test sample of 0.54 for the ASS-NN, against 0.49 for the linear MNL model.

Table 5 summarises the mean, bias and Root Mean Squared Error (RMSE) of the marginal utilities obtained with the ASS-NN, contrasted with the true values and with the values of a MNL model with a linear utility function. The mean of the marginal utility is presented since the ASS-NN predicts these values at the choice situation level. In contrast, the bias and RMSE are calculated to quantify the extent that the marginal utilities deviate from the true values.

Table 5: Marginal utilities of the ASS-NN against true values and from a linear MNL model

|  |  | Mean | | | Bias | | RMSE | |
|---|---|---|---|---|---|---|---|---|
|  |  | True value | Linear MNL | **ASS-NN** | Linear MNL | **ASS-NN** | Linear MNL | **ASS-NN** |
| Dataset 1 (Linear) | Train cost | -2 | -2.01 | **-2.00** | -0.01 | **< -0.01** | 0.01 | **0.07** |
|  | Train time | -3 | -3.05 | **-2.86** | -0.05 | **0.13** | 0.05 | **0.24** |
|  | SM cost | -2 | -2.01 | **-1.98** | -0.01 | **0.02** | 0.01 | **0.11** |
|  | SM time | -3 | -3.05 | **-2.90** | -0.05 | **0.09** | 0.05 | **0.19** |
|  | Car cost | -2 | -2.01 | **-2.01** | -0.01 | **-0.01** | 0.01 | **0.05** |
|  | Car time | -3 | -3.05 | **-2.92** | -0.05 | **0.08** | 0.05 | **0.14** |
| Dataset 2 (Log-linear) | Train cost | -5.53 | -2.14 | **-4.06** | 3.39 | **1.68** | 7.41 | **5.28** |
|  | Train time | -3.23 | -3.68 | **-3.03** | -0.44 | **0.20** | 1.54 | **0.59** |
|  | SM cost | -5.00 | -2.14 | **-3.60** | 2.87 | **1.60** | 7.21 | **5.26** |
|  | SM time | -6.04 | -3.68 | **-5.31** | 2.36 | **0.73** | 3.66 | **1.91** |
|  | Car cost | -3.59 | -2.14 | **-3.39** | 1.45 | **0.19** | 2.36 | **0.37** |
|  | Car time | -3.90 | -3.68 | **-3.57** | 0.22 | **0.32** | 1.90 | **0.97** |

For the linear-in-parameters utility DGP (dataset 1), the ASS-NN succeeds on recovering the true marginal utility travel cost on the average, with small numerical differences between the mean estimate and the true values. In contrast, for the marginal utility of travel time, these differences are higher. This can be explained by the selected structure of the ASS-NN. On the one hand, the cost-specific utility is modelled as a set of layers with shared weights across alternatives, which are trained with the attribute levels of all alternatives together. Thus, the cost-specific utility is modelled with a lower number of weights and a bigger amount of data than the other attributes. On the other hand, the travel time is modelled with independent sets of layers per alternative, with their own sets of independent weights, which are trained only with the data for its specific alternative.

For the log-linear utility DGP (dataset 2), the ASS-NN outperforms the MNL model with linear function on recovering all the predicted marginal utilities on average, compared with the true values. Furthermore, the ASS-NN consistently reaches lower average bias and RMSE than the linear MNL model. We also observe that the ASS-NN has a greater ability of recovering the marginal utility of travel time than the marginal utility of travel cost, as the bias and RMSE of the former are lower than the latter, in contrast to the findings of dataset 1, where the ASS-NN gets closer-to-truth marginal utilities of cost than marginal utilities of time.

Table 6 summarises the mean, bias and RMSE of the VTT for each mode, contrasted with the true values from the DGP and a linear MNL model. On average, the ASS-NN successfully recovers the true VTT per mode, with different degrees of precision per pseudo-synthetic dataset. On the one hand, when the DGP is a linear utility function (dataset 1), the mean bias of the VTTs obtained from the ASS-NN is between -0.15 and -0.07, while the RMSE values lie between 0.05 and 0.03. Both values are higher than those from a linear MNL model. On the other hand, when the true DGP is a log-linear function (dataset 2), the mean VTT values of the ASS-NN are closer to the true values compared with the VTTs obtained from the linear MNL model, which is also reflected in the differences of the bias and RMSE between each model. Furthermore, we observe higher precision of the ASS-NN on predicting VTTs than marginal utilities, as the bias and RMSE of the former are considerably smaller than those from the latter.

Table 6: VTT of the ASS-NN, compared with true values and a linear MNL model

|  |  | Mean | | | Bias | | RMSE | |
| --- | --- | --- | --- | --- | --- | --- | --- | --- |
|  |  | True value | Linear MNL | **ASS-NN** | Linear MNL | **ASS-NN** | Linear MNL | **ASS-NN** |
| Dataset 1 | Train | 1.50 | 1.52 | **1.42** | 0.02 | **-0.08** | 0.02 | **0.01** |
|  | Swissmetro | 1.50 | 1.52 | **1.45** | 0.02 | **-0.05** | 0.02 | **0.01** |
|  | Car | 1.50 | 1.52 | **1.44** | 0.02 | **-0.05** | 0.02 | **< 0.01** |
| Dataset 2 | Train | 0.98 | 1.72 | **0.85** | 0.73 | **0.01** | 1.14 | **0.02** |
|  | Swissmetro | 2.24 | 1.72 | **1.87** | -0.52 | **-0.06** | 2.11 | **0.19** |
|  | Car | 1.17 | 1.72 | **1.07** | 0.55 | **-0.03** | 0.72 | **0.04** |

## 4 Application with empirical data

The results of these Monte Carlo analyses show that the ASS-NN can learn the utility function from the data and provides interpretable outcomes close to the true values. In this section, we apply the ASS-NN to empirical choices from the Swissmetro data.

### 4.1 Data description

The data used in this section is the same as in the Monte Carlo analysis, i.e., the Swissmetro dataset, but includes the respondents' actual choices, instead of using pseudo-synthetic choices. The empirical dataset comprises 9,036 choice situations from 1,858 individuals, with their respective attribute levels and responses. Table 7 summarises the observed market shares, which evidence unbalance of the chosen transport modes. The most chosen mode is Swissmetro, with 5,177 observations (57.3% of the sample), followed by car with 3,080 observations (34.1% of the sample), and train with 779 times (8.6% of the sample).

Table 7: Observed market shares per travel mode

|  | Frequency | % |
| --- | --- | --- |
| Train | 779 | 8.6% |
| Car | 5,177 | 57.3% |
| Swissmetro | 3,080 | 34.1% |

We use the travel cost and travel time of each travel mode and the travel headway of train and Swissmetro as inputs of the ASS-NN. The travel cost variables of each mode are modelled with hidden layers with shared weights, while the travel time and headway are modelled with alternative-specific layers with independent weights. All attributes are scaled to hundreds (i.e., divided by 100) in order to avoid numerical overflow issues in the MNL models.

Table 8 presents the summary statistics of the (unscaled) attributes used for the analysis. On average and median, Swissmetro is the most expensive mode, but at the same time is the fastest in terms of travel time. The minimum travel cost of train and Swissmetro equals zero, as a difference from the minimum travel cost for car. This is because some respondents stated they own an annual public transport card that allows them to travel for free. The cost of such a card can be assumed as sunk costs (since travellers already paid for it), and therefore, the travel cost for such respondents is zero regardless of the trip they select as long as it is by public transport. In terms of travel time, Swissmetro is the fastest mode on average and in terms of mean values, followed by car and train.

Table 8: Summary statistics of the attributes per alternative

|  |  | Mean | Median | Std. Dev. | Minimum | Maximum |
|---|---|---|---|---|---|---|
| Cost (CHF) | Train | 91.27 | 81 | 65.51 | 0 | 576 |
|  | Swissmetro | 110.19 | 97 | 80.47 | 0 | 768 |
|  | Car | 93.43 | 84 | 47.48 | 8 | 520 |
| Time (minutes) | Train | 173.69 | 167 | 78.58 | 31 | 1,049 |
|  | Swissmetro | 91.38 | 81 | 55.76 | 8 | 796 |
|  | Car | 146.88 | 136 | 77.17 | 32 | 1,560 |
| Headway (mins/service) | Train | 70.04 | 60 | 37.42 | 30 | 120 |
|  | Swissmetro | 20.05 | 20 | 8.16 | 10 | 30 |

Figure 2 shows the distribution of the (unscaled) travel time and travel costs per mode, which were designed following a pivoted design, i.e., based on the respondents' current travel time and travel cost. The travel headway, on the other hand, was designed based on three possible levels per mode: the trains' headway could be either 30, 60 or 120 minutes per train, whereas the headway of Swissmetro could be either 10, 20 or 30 minutes per service. The distributions of travel cost and travel time per mode are skewed toward the left. Most respondents faced travel costs between zero and 400 CHF, while higher values can be considered outliers. A considerable part of respondents has travel costs equal to zero for train and Swissmetro, which are those who own the annual public transport card. Regarding travel time, most respondents faced between zero and 4 hours (approximately 200 minutes), with slightly longer travel times for train and car trips, compared with Swissmetro trips.

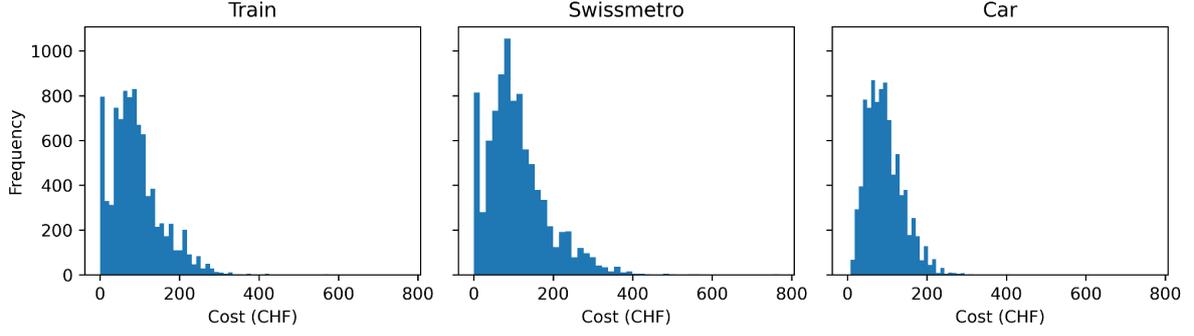

(a) travel cost

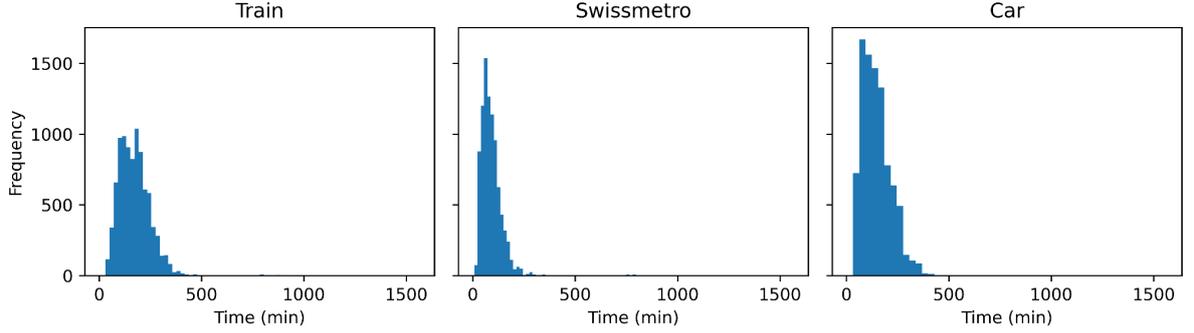

(b) travel time

Figure 2: Distribution of travel cost and travel time per mode

## *4.2 MNL models for contrast*

In addition to comparing with the results of the ASU-DNN, we contrast the results of the ASS-NN model with two MNL models with different utility function specifications. The first model is specified with a linear-in-parameters utility function (henceforth a linear MNL model), in which the cost-specific parameter is equal across alternatives, whereas the parameters of travel time and headway are alternative-specific. Additionally, we include alternative-specific constants to reflect the labelled nature of the choice experiment. Thus, the (observed) utility of this model is defined as in equations (10) to (12):

$$V_{TRAIN} = \beta_{TC} \cdot TC_{TRAIN} + \beta_{TT,TRAIN} \cdot TT_{TRAIN} + \beta_{HE,TRAIN} \cdot HE_{TRAIN}, \quad (10)$$

$$V_{SM} = \alpha_{SM} + \beta_{TC} \cdot TC_{SM} + \beta_{TT,SM} \cdot TT_{SM} + \beta_{HE,SM} \cdot HE_{SM}, \quad (11)$$

$$V_{CAR} = \alpha_{CAR} + \beta_{TC} \cdot TC_{CAR} + \beta_{TT,CAR} \cdot TT_{CAR}, \quad (12)$$

where $\alpha_{SM}$ and $\alpha_{CAR}$ are alternative-specific constants, $\beta_{TC}$ is the alternative-shared parameter associated with travel cost, $\beta_{TT,j}$ are the alternative-specific parameters associated with travel time, and $\beta_{HE,j}$ are the alternative-specific parameters associated with headway.

The second model is a MNL model with a log-linear utility function (henceforth the log-linear MNL model), which is formulated in a similar way as in the linear MNL model, but with attributes in logarithms. To address numerical overflow when the travel cost is equal to zero, we add a constant to all attributes equal to 0.1. The utility functions are specified as in equations (13) to (15):

$$V_{TRAIN} = \beta_{TC} \cdot \ln(TC_{TRAIN} + 0.1) + \beta_{TT,TRAIN} \cdot \ln(TT_{TRAIN} + 0.1) + \beta_{HE,TRAIN} \quad (13)$$
$$\cdot \ln(HE_{TRAIN} + 0.1),$$

$$V_{SM} = \alpha_{SM} + \beta_{TC} \cdot \ln(TC_{SM} + 0.1) + \beta_{TT,SM} \cdot \ln(TT_{SM} + 0.1) + \beta_{HE,SM} \quad (14)$$
$$\cdot \ln(HE_{SM} + 0.1),$$

$$V_{CAR} = \alpha_{CAR} + \beta_{TC} \cdot \ln(TC_{CAR} + 0.1) + \beta_{TT,CAR} \cdot \ln(TT_{CAR} + 0.1), \quad (15)$$

## 4.3 Results with empirical data

### 4.3.1 Goodness-of-fit and estimation/training time

Table 9 summarises the training results of the ASS-NN, contrasted with the ASU-DNN and the MNL models. The ASU-DNN reaches the highest log-likelihood in all samples and Rho-squared in the test sample, followed by the ASS-NN. This result is expected since the ASS-NN is a restricted version of the ASU-DNN. Compared with the MNL models, the ASS-NN outperforms both linear and log-linear MNL models in terms of Rho-squared (test sample): the difference between the ASS-NN and the linear MNL model is of 0.3 points and 0.1 points for the log-linear MNL model. These results sign potential non-linear effects in the underlying DGP that a linear MNL models does not account for, whereas they are captured to a greater extent by the log-linear MNL, the ASU-DNN and the ASS-NN. In terms of estimation/training time, the MNL models are considerably faster, with less than one second of estimation time, while the ASS-NN and the ASU-DNN take between 10 and 16 seconds per training repetition, on average.

Table 9: Training results of the ASS-NN compared with the ASU-DNN and MNL models, empirical data.

|  | **ASS-NN** | ASU-DNN | Linear MNL | Log-linear MNL |
| --- | --- | --- | --- | --- |
| Log-likelihood (full sample) | **-6,940.79** | -6,747.95 | -7,214.94 | -6,994.60 |
| Log-likelihood (training sample) | **-5,548.44** | -5,388.77 | -5,767.31 | -5,591.37 |
| Log-likelihood (test sample) | **-1,392.35** | -1,359.18 | -1,447.63 | -1,403.24 |
| Rho-squared (test sample) | **0.30** | 0.32 | 0.27 | 0.29 |
| Estimation/training time (secs.) | **9.98s./training** | 16.24s./training | < 1sec. (total) | < 1sec. (total) |
| Optimal number of hidden layers | **2 layers** | 2 layers | - | - |
| Optimal number of nodes per layer | **10 nodes** | 10 nodes | - | - |
| Activation function | **tanh** | tanh | - | - |

### 4.3.2 Marginal utility of attribute increases

Table 10 presents the average marginal utilities obtained with the ASS-NN, compared with the predictions of the ASU-DNN and MNL models. The ASS-NN predicts an average marginal utility of cost that slightly varies across modes (between -1.51 and -1.34), which is explained by the differences in travel costs presented in the choice experiment (see summary statistics of Table 9). In contrast, the ASU-DNN predicts considerably different marginal utilities of cost per mode (between -2.56 and - 0.86). This is expected, as the ASU-DNN does not restrict the cost-dependent utilities to have the same form across different alternatives. Compared to the MNL models, the ASS-NN predicts a higher marginal utility of costs than the log-linear MNL model (between -2.03 and -1.28). Finally, the linear MNL model predicts a constant marginal utility of cost of -0.81.

Table 10: Average predicted marginal utilities ASS-NN, ASU-DNN and MNL models. Empirical data.

| Mode | Attribute | **ASS-NN** | ASU-DNN | Linear MNL | Log-linear MNL |
| --- | --- | --- | --- | --- | --- |
| Train | Cost (x100) | **-1.51** | -2.56 | -0.81 | -2.03 |
|  | Time (x100) | **-2.06** | -1.80 | -2.04 | -2.28 |
|  | Headway (x100) | **-0.88** | -0.91 | -0.78 | -0.92 |
| Swissmetro | Cost (x100) | **-1.34** | -1.43 | -0.81 | -1.85 |
|  | Time (x100) | **-2.06** | -2.06 | -1.51 | -2.34 |
|  | Headway (x100) | **-1.12** | -1.19 | -0.73 | -0.71 |
| Car | Cost (x100) | **-1.48** | -0.86 | -0.81 | -1.28 |
|  | Time (x100) | **-1.10** | -1.46 | -1.00 | -1.30 |

In terms of travel time, the ASS-NN predicts a higher average marginal utility for train and Swissmetro trips (-2.06) than for car trips (-1.10), which signs that individuals have a higher sensitivity for saving travel time for average train and Swissmetro trip than for the average car trip. The ASU-DNN follows a similar pattern, with higher marginal utility of time for train and Swissmetro trips (-1.80 and -2.06, respectively), compared with car trips (-1.46). Compared with the MNL models, we observe that the ASS-NN is more conservative than a log-linear MNL model regarding the average marginal utility of travel time of train and Swissmetro trips, while for car trips, the marginal utility of time is the same. In contrast, compared with the linear MNL model, the ASS-NN predicts a higher average marginal utility of travel time for Swissmetro and car trips. Finally, the ASS-NN predicts a higher predicted marginal utility of headway for Swissmetro trips (-1.12) than train trips (-0.88). In contrast, the MNL models predict higher values for train than for Swissmetro trips.

To explore differences in the marginal utility for different attribute levels, we plot the predicted marginal utilities of the ASS-NN for each mode against their respective attribute values. These plots are illustrated in Figure 3. As expected, the predicted marginal utility of costs is equal for the same cost levels across different travel modes due to the consistency of the ASS-NN with the fungibility of money assumption. Conversely, the ASS-NN predicts different shapes of the marginal utility of travel time and headway per travel mode at different values of the associated levels, respectively. For travel time, we observe that the marginal utility of car trips is more inelastic than train and Swissmetro trips. Furthermore, the predicted marginal utilities of travel time suggest that, for trips up to 180 minutes (3 hours) approximately, individuals are more sensitive to changes in travel time for train and Swissmetro trips than for car trips, *ceteris paribus*. For trips between 180 and 300 minutes (3 to 5 hours) approximately, respondents are more sensitive to changes in travel time for train and car trips than for Swissmetro trips, *ceteris paribus.* For trips of 300 minutes (5 hours) approximately or more, individuals are more sensitive to travel time changes of car trips than the other modes. Finally, we observe that the slope of the marginal utility of travel headway for Swissmetro trips is steeper than for train trips as the waiting time per service increases. This suggests that, in terms of travel headway, the marginal utility for Swissmetro trips is more elastic than for train trips.

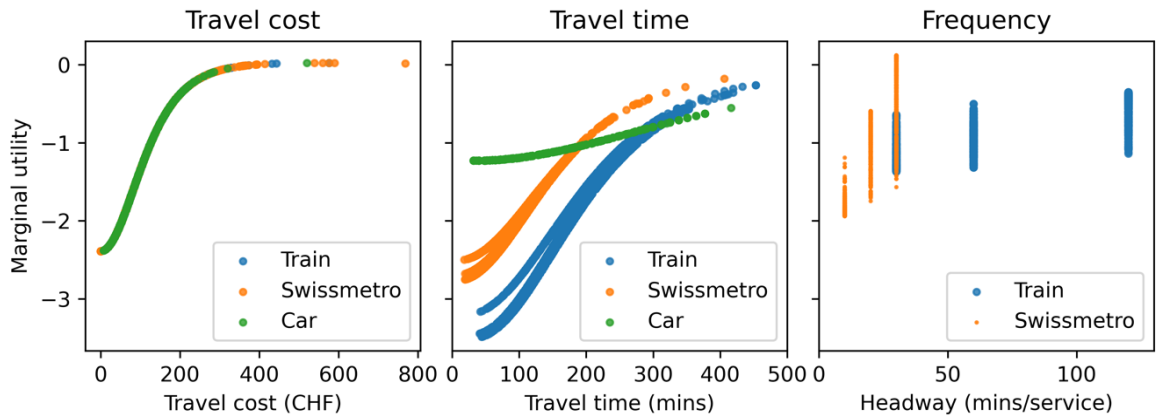

Figure 3: Predicted marginal utility with respect to travel time, cost and headway, empirical data.

### 4.3.3 VTT and VoWT

Table 11 summarises the average predicted VTT and VoWT by the ASS-NN per travel mode, contrasted with the predictions of the ASU-DNN and MNL models. The VTT and VoWT values presented in this table are computed after dropping outliers (upper 5% of the sample), negative VTT values (up to 16 observations) and negative VoWT values (up to 105 observations).

Table 11: Average predicted VTT and VoWT compared with the ASU-DNN and MNL models, empirical data.

|  | VTT (CHF/min) | | | | VoWT (CHF/min) | | | |
| --- | --- | --- | --- | --- | --- | --- | --- | --- |
|  | **ASS-NN** | ASU-DNN | Linear MNL | Log-linear MNL | **ASS-NN** | ASU-DNN | Linear MNL | Log-linear MNL |
| Train | **1.52** | 0.69 | 2.52 | 1.73 | **0.68** | 0.34 | 0.96 | 0.78 |
| Swissmetro | **2.11** | 1.51 | 1.86 | 2.13 | **1.14** | 0.90 | 0.90 | 0.72 |
| Car | **0.81** | 1.70 | 1.24 | 1.03 | - |  | - | - |

The ASS-NN predicts that Swissmetro trips have the highest average VTT (2.11 CHF/min), followed by train trips (1.52 CHF/min) and car trips (0.81 CHF/min). This order pattern is similar to the log-linear MNL model, whereas the linear MNL model predicts the highest average VTT for train trips, followed by Swissmetro and car trips. In contrast, the ASU-DNN predicts a higher average VTT for car trips, followed by Swissmetro and train trips. In magnitude, the ASS-NN consistently predicts more conservative average VTT values than the MNL models, except in the case of Swissmetro trips, where the lowest predicted VTT is in the linear MNL model. Compared to the ASU-DNN, the ASS-NN predicts a higher VTT except for car trips. In terms of the VoWT, the ASS-NN predicts the highest value for Swissmetro trips (1.14 CHF/min), followed by train trips (0.68 CHF/min). The same pattern in followed in the ASU-DNN, where Swissmetro trips have the highest VoWT (0.9 CHF/min), followed by train trips (0.34 CHF/min). In contrast, in the MNL models, the highest predicted VoWT is associated with train trips. However, the differences across different modes for the same model are not substantial in magnitude in the MNL models. Furthermore, it is more reasonable that the VoWT of Swissmetro trips would be higher than for train trips, as the former trips have a higher frequency and are more expensive than the latter trips.

Finally, Figure 4 compares the average VTT of the ASS-NN per travel mode for different trip times. We observe that for short trips (less than 60 minutes), the average VTT of train and Swissmetro trips are similar and close to 2 CHF/min, whereas the average VTT for car trips is considerably lower (0.5 CHF/min approx.). For trips between 60 and 89 minutes (1 to 1.5 hours), the average VTT of Swissmetro trips rises considerably to reach almost 2.5 CHF/min, while it decreases for train trips to 1.5 CHF/min, which evidences a potential mode switch from train to Swissmetro for trips of this time range. As the travel time increases, the average VTT of Swissmetro trips decreases, to the extent of intersecting the average VTT of train trips in the group of 180 to 239 minutes (3 to 4 hours), while the average VTT of car trips steadily increases. We can expect that, for longer trips, the average VTT of a car would surpass the average VTT of train and Swissmetro, if these results follow the same trajectory, which can be evidence that, for long trips, individuals would switch to a car for such travel.

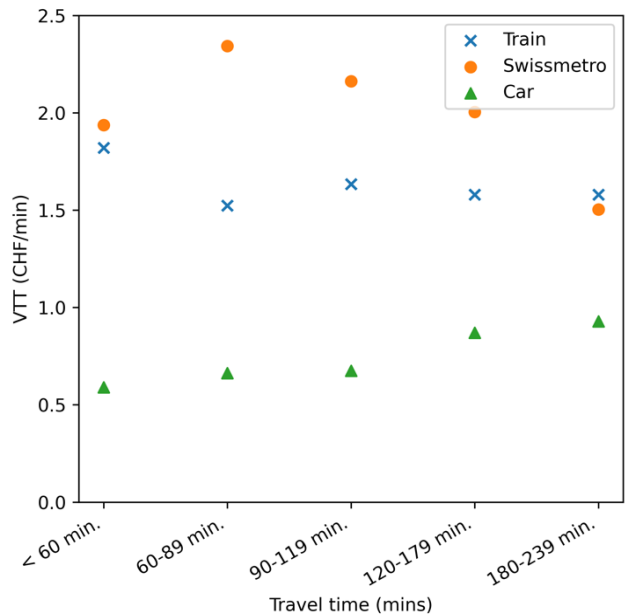

Figure 4: average VTT per mode for different travel time ranges.

## 5   Discussion and conclusion

In this paper, we propose a new discrete choice model based on ANNs, called ASS-NN. The ASS-NN is based in the ASU-DNN, which balances a flexible utility approximation from the data and satisfies consistency with RUM theory. In addition, the ASS-NN is consistent with the assumption of fungibility of money (i.e., one euro is one euro). By accommodating for these assumptions, the ASS-NN considers that money can be spent in different goods interchangeably, which is key for deriving economically-sound welfare measures.

*5.1 Main findings*

The Monte Carlo experiment shows that the ASS-NN succeeds on approximating the utility from the data under different model specifications, reaching goodness of fit (i.e., log-likelihood and Rho-squared) close to the ground truth. Furthermore, we show that the ASS-NN has a higher accuracy on recovering the marginal utility of attribute increases and VTT than a misspecified MNL model. The differences between the recovered marginal utilities and VTT across different utility specifications are in line with previous findings in the literature (Torres et al., 2011; van der Pol et al., 2014). Our findings in the Monte Carlo experiment support the use of the ASS-NN for recovering interpretable outcomes without the need of explicitly defining the utility's functional form.

Our empirical results show that, without specifying the utility functional form, the ASS-NN outperforms (in terms of goodness of fit) MNL models under two different utility specifications widely used in other empirical studies, namely linear and log-linear utility functions. Furthermore, the ASS-NN predicts a marginal utility of costs consistent with the fungibility of money assumption (see Figure 3), as a difference with the ASU-DNN. Regarding the marginal utility of travel time, the ASS-NN predicts differences across different modes for the same travel time value, with public transport modes being more attractive for short-distance trips. At the same time, this trend reverts for long-distance trips.

Furthermore, we found that respondents assign a higher average value of travel time (VTT) to public transport trips, particularly train and Swissmetro trips, than car trips. As the trip length increases, the VTT for train and Swissmetro trips decreases in favour of car trips. The study also shows a higher average value of waiting time (VoWT) for Swissmetro trips compared to train trips, which contradicts the predictions of the MNL models.

*5.2 Limitations and further research directions*

While we identify promising implications and uses of the ASS-NN, we also acknowledge three limitations of our work. Firstly, ANNs are known for requiring a higher amount of data than DCMs. As shown by Alwosheel et al. (2018), ANNs require around 50 times the amount of data per estimated weight, much higher than in most conventional choice models, including the MNL model. Such a criterion applied in the context of our paper implies that the number of weights of each ASS-NN should be around 180. A second limitation of our work is the treatment of unreasonably high or below zero VTT/VoWT values. The former case is a consequence of marginal utility values of cost that lie close to zero (Sillano & de Dios Ortúzar, 2005), while the latter is not theoretically possible from an economic perspective (Hess et al., 2005). We relied on dropping such problematic VTT/VoWT values from the sample before presenting the results. However, as such values still may provide relevant behavioural information, further research should be done to treat these cases more elegantly and properly. Finally, the *Swissmetro* dataset could be rather small for the standard practice of ANN-based models, which can explain the small goodness-of-fit improvements in our applications.

We envision three further research direction from this work. First, we foresee the possibility of exploring methodologies to incorporate panel choices and random parameters to the ASS-NN, equivalent to Mixed Logit models. Panel structures can be incorporated by slight modifications in the network architecture of the ASS-NN. At the same time, random parameters may require alternative network structures that account for probability distributions, such as probabilistic neural networks (Mao et al., 2000). Secondly, we suggest testing the role of sample size on the results of the ASS-NN, for instance, with bigger datasets such as the London Passenger Mode Choice data (Hillel et al. 2018). Thirdly, the envision an extension of the ASS-NN to incorporate sociodemographic characteristics, in a similar way of the work of Sifringer et al. (2020). Overall, we observe that the flexibility of ANNs provides clear opportunities further to incorporate machine learning methodologies in our choice modelling toolbox.

**Acknowledgements**

The authors acknowledge the support of the TU Delft AI Labs programme. José Ignacio Hernández would like to acknowledge the Netherlands Organisation for Scientific Research (NWO) for funding his PhD (NWO Responsible Innovation grant – 313-99-333) and the valuable feedback and suggestions from Francisco Garrido Valenzuela during the conception and development of this work. Also, we are thankful for the insightful feedback provided by three anonymous reviewers during the revision stage of this paper.